\documentclass[letterpaper, 10 pt, conference]{ieeeconf}  

\IEEEoverridecommandlockouts                              

\usepackage{xcolor}

\usepackage{graphicx}

\usepackage{gensymb}
\usepackage[hyphens]{url}
\usepackage{multirow}
\usepackage{multicol}
\usepackage{tabularx}
\usepackage{dsfont}
\usepackage{url}
\usepackage{color}
\usepackage{subcaption}
\usepackage{caption}
\usepackage{booktabs}
\usepackage{dcolumn}
\usepackage{amssymb}
\usepackage{hyperref}

\hypersetup{
    colorlinks=false,
    linkcolor=blue,
    filecolor=magenta,      
    urlcolor=cyan,
    pdftitle={Overleaf Example},
    pdfpagemode=FullScreen,
    }

\newcommand{\etal}{{et al}.\@ }
\title{\LARGE \bf
OPV2V: An Open Benchmark Dataset and Fusion Pipeline for Perception with Vehicle-to-Vehicle Communication}

\author{Runsheng Xu$^{1*}$, Hao Xiang$^{1*}$, Xin Xia$^{1}$, Xu Han$^{1}$, Jinlong Li$^{2}$, Jiaqi Ma$^{1}$

\thanks{*Equal contribution}
\thanks{$^{1}$University of California, Los Angeles, Mobility Lab. {\{\tt\small{rxx3386, haxiang, x35xia, hanxu417, jiaqima}\}}{\tt\small{@ucla.edu}}}%
\thanks{$^{2}$Cleveland State University, Cleveland Vision and AI Lab,
{\tt\small j.li56@vikes.csuohio.edu}}%
}

\begin{document}

 \maketitle
\thispagestyle{empty}
\pagestyle{empty}

\begin{abstract}

Employing Vehicle-to-Vehicle communication to enhance perception performance in self-driving technology has attracted considerable attention recently; however, the absence of a suitable open dataset for benchmarking algorithms has made it difficult to develop and assess cooperative perception technologies. To this end, we present the first large-scale open simulated dataset for Vehicle-to-Vehicle perception. It contains over 70 interesting scenes, 11,464 frames, and 232,913 annotated 3D vehicle bounding boxes, collected from 8 towns in CARLA and a digital town of Culver City, Los Angeles. We then construct a comprehensive benchmark with a total of 16 implemented models to evaluate several information fusion strategies~(i.e. early, late, and intermediate fusion) with state-of-the-art LiDAR detection algorithms. Moreover, we propose a new Attentive Intermediate Fusion pipeline to aggregate information from multiple connected vehicles. Our experiments show that the proposed pipeline can be easily integrated with existing 3D LiDAR detectors and achieve outstanding performance even with large compression rates. To encourage more researchers to investigate Vehicle-to-Vehicle perception, we will release the dataset, benchmark methods, and all related codes in 
\textcolor{blue}{\href{https://mobility-lab.seas.ucla.edu/opv2v/}{https://mobility-lab.seas.ucla.edu/opv2v/}}.



\end{abstract}

\section{INTRODUCTION}

Perceiving the dynamic environment accurately is critical for robust intelligent driving. With recent advancements in robotic sensing and machine learning, the reliability of perception has been significantly improved~\cite{liu2019auto, he2016deep, qi2017pointnet}, and 3D object detection algorithms have achieved outstanding performance either with LiDAR point clouds~\cite{voxelnet,pointpillar,Shi_2019_CVPR, Shi2020PVRCNNPF} or multi-sensor data~\cite{Liang2009, Liang2018}. 

\begin{figure*}[!t]
\centering
\begin{subfigure}[c]{0.9\linewidth}
    \centering{\includegraphics[width=1\linewidth]{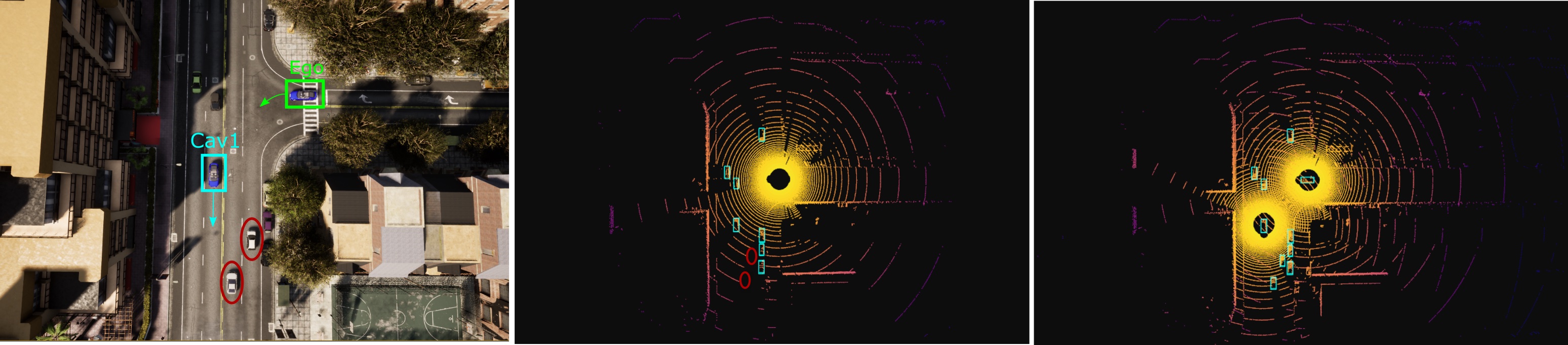}}
    \caption{}
    \label{fig:overview-a}
\end{subfigure}
\begin{subfigure}[c]{0.9\linewidth}
    \centering{\includegraphics[width=1\linewidth]{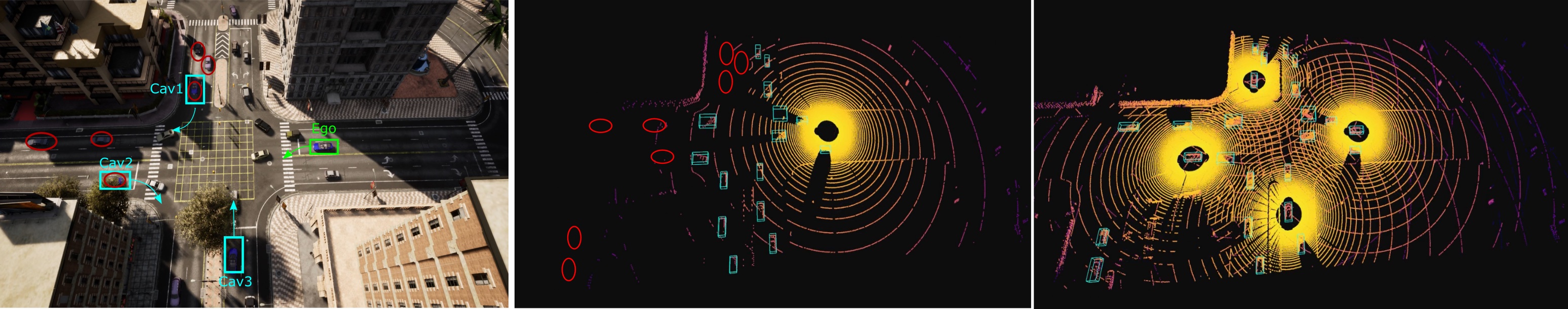}}
    \caption{}
\end{subfigure}
\caption{Two examples from our dataset. \emph{Left}: Screenshot of the constructed scenarios in CARLA. \emph{Middle}: The LiDAR point cloud collected by the ego vehicle. \emph{Right}: The aggregated point clouds from all surrounding CAVs. The red circles represent the cars that are invisible to the ego vehicle due to the occlusion but can be seen by other connected vehicles. (a): The ego vehicle plans to turn left in a T-intersection and the roadside vehicles block its sight to the incoming traffic. (b): Ego-vehicle's LiDAR has no measurements on several cars because of the occlusion caused by the dense traffic.}
\label{fig:overview}
\end{figure*}

Despite the recent breakthroughs in the perception field, challenges remain. When the objects are heavily occluded or have small scales, the detection performance will dramatically drop. Such problems can lead to catastrophic accidents and are difficult to solve by any algorithms since the sensor observations are too sparse. An example is revealed in Fig.~\ref{fig:overview-a}. Such circumstances are very common but dangerous in real-world scenarios, and these blind spot issues are extremely tough to handle by a single self-driving car.

To this end, researchers started recently investigating dynamic agent detection in a cooperative fashion, such as USDOT CARMA~\cite{lochrane2020carma} and Cooper~\cite{cooper}. By leveraging the Vehicle-to-Vehicle~(V2V) communication technology, different Connected Automated Vehicles~(CAVs) can share their sensing information and thus provide multiple viewpoints for the same obstacle to compensate each other. The shared information could be raw data, intermediate features, single CAV's detection output, and metadata e.g., timestamps and poses. Despite the big potential in this field, it is still in its infancy. One of the major barriers is the lack of a large open-source dataset. Unlike the single vehicle's perception area where multiple large-scale public datasets exist~\cite{nuscenes2019, sun2020scalability, Geiger2013IJRR}, most of the current V2V perception algorithms conduct experiments based on their customized data~\cite{Wang2020V2VNetVC, rawaw018, Zhang2021DistributedDM}. These datasets are either too small in scale and variance or they are not publicly available. Consequently, there is no large-scale dataset suitable for benchmarking distinct V2V perception algorithms, and such deficiency will preclude further progress in this research field. 

\begin{figure}
    \centering
    \begin{subfigure}[c]{1\linewidth}
        \centering\includegraphics[width=1\linewidth]{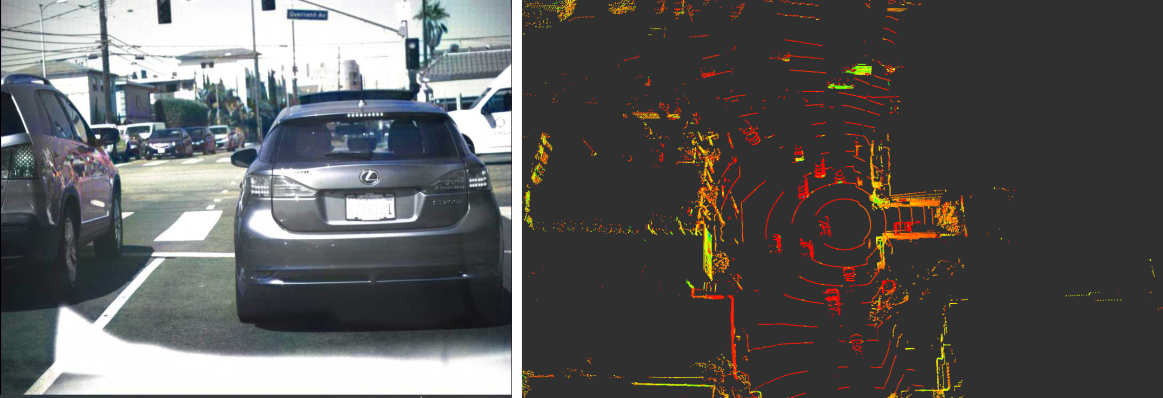}
        \caption{}
    \end{subfigure}
    \begin{subfigure}[c]{1\linewidth}
        \centering\includegraphics[width=1\linewidth]{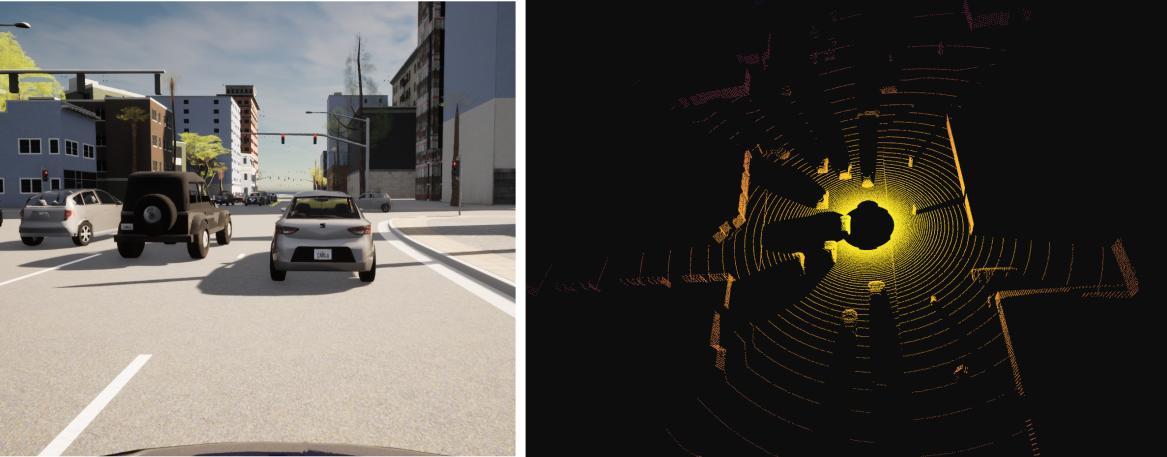}
        \caption{}
    \end{subfigure}
    \caption{A caparison between the real Culver City and its digital town. (a) The RGB image and LiDAR point cloud captured by our vehicle in Culver City. (b) The corresponding frame in the digital town. The road topology, building layout, and traffic distribution are similar to reality.  }
    \label{fig:culver}
\end{figure}

To address this gap, we present OPV2V, the first large-scale \textbf{O}pen Dataset for \textbf{P}erception with \textbf{V2V} communication. By utilizing a cooperative driving co-simulation framework named OpenCDA~\cite{xu2021opencda} and CARLA simulator~\cite{Dosovitskiy17}, we collect 73 divergent scenes with a various number of connected vehicles to cover challenging driving situations like severe occlusions. To narrow down the gap between the simulation and real-world traffic, we further build a digital town of Culver City, Los Angeles with the same road topology and spawn dynamic agents that mimic the realistic traffic flow on it.  Data samples are shown in Fig.~\ref{fig:overview} and Fig.~\ref{fig:culver}. We benchmark several state-of-the-art 3D object detection algorithms combined with different multi-vehicle fusion strategies. On top of that, we propose an Attentive Intermediate Fusion pipeline to better capture interactions between connected agents within the network. Our experiments show that the proposed pipeline can efficiently reduce the bandwidth requirements while achieving state-of-the-art performance.

\section{Related Work}
\noindent\textbf{Vehicle-to-Vehicle Perception: }V2V perception methods can be divided into three categories: early fusion, late fusion, and intermediate fusion. Early fusion methods~\cite{cooper} share raw data with CAVs within the communication range, and the ego vehicle will predict the objects based on the aggregated data. These methods preserve the complete sensor measurements but require large bandwidth and are hard to operate in real time~\cite{Wang2020V2VNetVC}. In contrast, late fusion methods transmit the detection outputs and fuse received proposals into a consistent prediction. Following this idea, Rauch \etal~\cite{Rauch2012} propose a Car2X-based perception module to jointly align the shared bounding box proposals spatially and temporally via an EKF. In~\cite{Rawashdeh2018}, a machine learning-based method is utilized to fuse proposals generated by different connected agents. This stream of work requires less bandwidth, but the performance of the model is highly dependent on each agent's performance within the vehicular network. To meet requirements of both bandwidth and detection accuracy, intermediate fusion~\cite{f-cooper, Wang2020V2VNetVC} has been investigated, where intermediate features are shared among connected vehicles and fused to infer the surrounding objects. F-Cooper~\cite{f-cooper} utilizes max pooling to aggregate shared Voxel features, and V2VNet~\cite{Wang2020V2VNetVC} jointly reason the bounding boxes and trajectories based on shared messages. 



\begin{table}[]
\centering
\begin{tabular}{l|l}
Sensors    & Details                                \\ \hline
4x Camera  & RGB, $800 \times 600$ resolution, $110\degree$ FOV     \\ \hline
1x LiDAR & \begin{tabular}[c]{@{}l@{}}$64$~channels, $1.3~M$ points per second, \\ $120~m$ capturaing range, $-25\degree$ to $5\degree$ \\ vertical FOV, $\pm2~cm$ error\end{tabular} \\ \hline
GPS \& IMU & $20~mm$ positional error, $2\degree$ heading error \\ \hline
\end{tabular}
\caption{Sensor specifications.}
\label{table:sensor-details}

\end{table}

\begin{figure}
    \centering
    \includegraphics[width=2.8in]{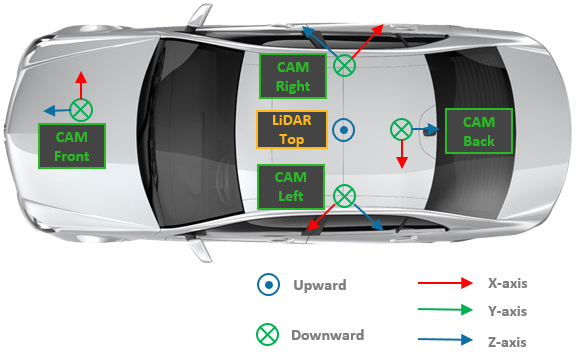}
    \caption{Sensor setup for each CAV in OPV2V. }
    \label{fig:sensor_config}
\end{figure}

\begin{figure}
    \centering
    \includegraphics[width=3.5in]{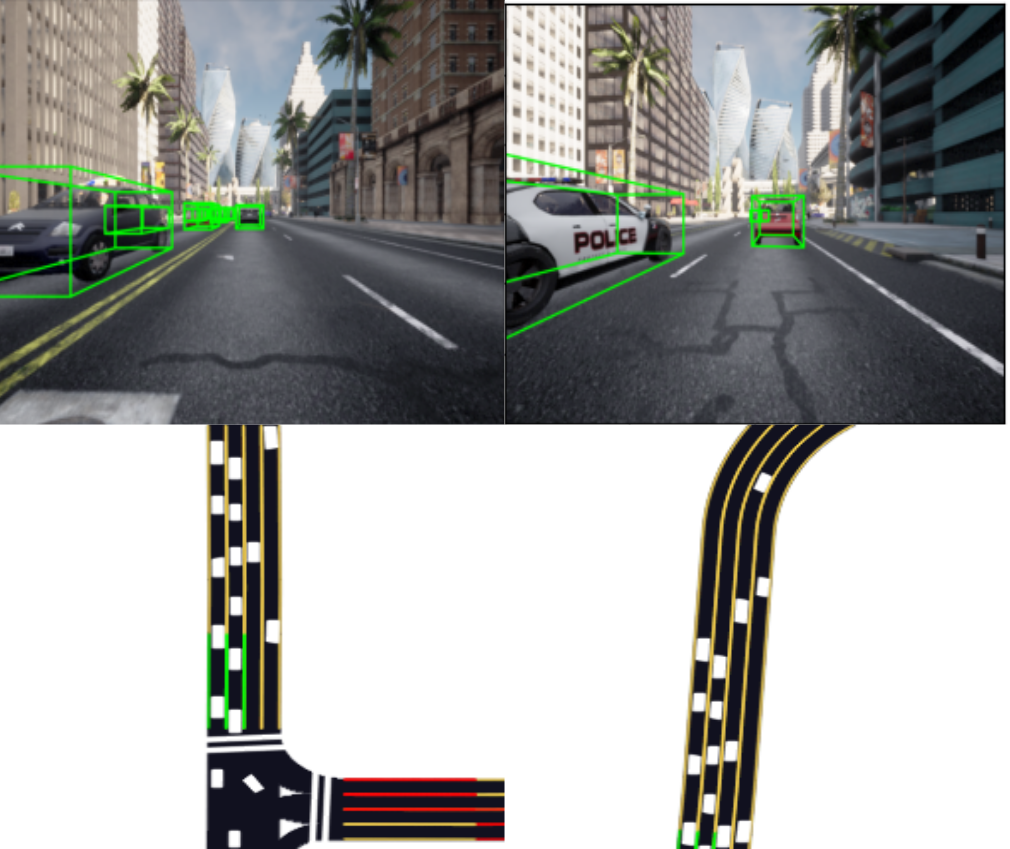}
    \caption{Examples of the front camera data and BEV map of two CAVs in OPV2V. The yellow, green, red, and white lanes in the BEV map represent the lanes without traffic light control, under green light control, under red light control, and crosswalks.}
    \label{fig:camera_example}
\end{figure}

\noindent\textbf{Vehicle-to-Vehicle Dataset: }To the best of our knowledge, there is no large-scale open-source dataset for V2V perception in the literature. Some work~\cite{cooper, f-cooper} adapts KITTI~\cite{Geiger2013IJRR} to emulate V2V settings by regarding the ego vehicle at different timestamps as multiple CAVs. Such synthetic procedure is unrealistic and not appropriate for V2V tasks since the dynamic agents will appear at different locations, leading to spatial and temporal inconsistency. \cite{Wang2020V2VNetVC} utilizes a high-fidelity LiDAR simulator~\cite{Lidarsim} to generate a large-scale V2V dataset. However, neither the LiDAR simulator nor the dataset is publicly available. Recently, several works~\cite{Zhang2021DistributedDM, marvasti2020cooperative} manage to evaluate their V2V perception algorithms on the CARLA simulator, but the collected data has a limited size and is restricted to a small area with a fixed number of connected vehicles. More importantly, their dataset is not released and difficult to reproduce the identical data based on their generation approach. T\&J dataset~\cite{cooper, f-cooper} utilizes two golf carts equipped with 16-channel LiDAR for data collection. Nevertheless, the released version only has 100 frames without ground truth labels and only covers a restricted number of road types. A comparison to existing dataset is provided in Table~\ref{table:dataset-comparison}.

\begin{table*}[]
    \centering
    \caption{Dataset comparison. ($^\dagger$) The number is reported based on data used during their experiment.
    ($^{\dagger\dagger}$) Single LiDAR resolution's data is counted.
    ($^\ddagger$) Ground truth data is not released in the T\&J dataset and it only has 100 frames and LiDAR data. (-) means that the number is not reported in the paper and can't be found in open dataset. ($^*$) means the data has the format mean$\pm$std.}
    \begin{tabular}{c|ccccc|ccc}
    \toprule
         \multirow{2}{*}{Dataset}  &  \multirow{2}{*}{frames}&  GT  &  Dataset  &  CAV  &  cities  &  Code  &  Open Dataset  &  Reproducibility\& \\
         &    &    3D boxes  &  Size  &  range  &    &    &    &Extensibility\\
         \hline
         V2V-Sim~\cite{Wang2020V2VNetVC}&51,200&-&-&10 $\pm$ 7$^*$&  $>1$  & &&  \\
         \cite{Zhang2021DistributedDM}  &  1,310$^\dagger$  &  -  &  -  &     3, 5 &  1  &    &    &\\
         \cite{marvasti2020cooperative}  &  6,000$^{\dagger\dagger}$  &  -  &  -  &  2  & 1  &  \checkmark  &    &     \\
         T\&J~\cite{cooper, f-cooper}  &  100$^\ddagger$  &  0$^\ddagger$  &  183.7MB  &  2  &  1  &  \checkmark  &  \checkmark \\ 
         OPV2V&11,464 & 232,913 &249.4GB &2.89 $\pm$ 1.06$^*$& 9&\checkmark&\checkmark&\checkmark\\ 
    \bottomrule
    \end{tabular}
    \label{table:dataset-comparison}
\end{table*}

\section{OPV2V Dataset}
\subsection{Data Collection}
\noindent\textbf{Simulator Selection.} CARLA is selected as our simulator to collect the dataset, but CARLA itself doesn't have V2V communication and cooperative driving functionalities by default. Hence, we employ OpenCDA~\cite{xu2021opencda}, a co-simulation tool integrated with CARLA and SUMO~\cite{OlaverriMonreal2018ConnectionOT}, to generate our dataset\footnote{Codes for generating our dataset have been recently released in https://github.com/ucla-mobility/OpenCDA/tree/feature/data\_collection}. It is featured with easy control of multiple CAVs, embedded vehicle network communication protocols, and more convenient and realistic traffic management.  

\noindent\textbf{Sensor Configuration.} The majority of our data comes from eight default towns provided by CARLA.  Our dataset has on average approximately 3 connected vehicles with a minimum of 2 and a maximum of 7 in each frame. As Fig.~\ref{fig:sensor_config} shows, each CAV is equipped with 4 cameras that can cover 360\degree~view together, a 64-channel LiDAR, and GPS/IMU sensors.  The sensor data is streamed at 20 Hz and recorded at 10 Hz. A more detailed description of the sensor configurations is depicted in Table~\ref{table:sensor-details}.


\noindent\textbf{Culver City Digital Town.} To incorporate scenarios that can better imitate real-world challenging driving environments and evaluate models' domain adaptation capability, we further gather several scenes imitating realistic configurations. An automated vehicle equipped with a 32-channel LiDAR and two cameras is sent out to Culver City during rush hour to collect sensing data. Then, we populate the road topology of digital town via RoadRunner~\cite{roadrunner}, select buildings based on agreement with collected data, and then spawn cars mimicking the real-world traffic flow with the support of OpenCDA. We collect 4 scenes in Culver City with around 600 frames in total (See Fig.~\ref{fig:culver}). These scenes will be used for validation of models trained with simulated datasets purely generated in CARLA. Future addition of data from real environments is planned and can be added to the model training set.

\noindent\textbf{Data Size.} Overall, 11,464 frames~(i.e. time steps) of LiDAR point clouds~(see Fig.~\ref{fig:overview}) and RGB images~(see Fig.~\ref{fig:camera_example}) are collected with a total file size of 249.4 GB. Moreover, we also generate Bird Eye View~(BEV) maps for each CAV in each frame to facilitate the fundamental BEV semantic segmentation task.

\noindent\textbf{Downstream Tasks.} By default, OPV2V supports cooperative 3D object detection, BEV semantic segmentation, tracking, and prediction either employing camera rigs or LiDAR sensors. To enable users to extend the initial data, we also provide a driving log replay tool\footnote{The tool can be found  \textcolor{blue}{\href{https://github.com/DerrickXuNu/OpenCOOD/tree/feature/log_replay/logreplay}{here}}.}.  along with the dataset. By utilizing this tool, users can define their own tasks~(e.g., depth estimation, sensor fusion) and set up additional sensors~(e.g., depth camera) without changing any original driving events. Note that in this paper, we only report the benchmark results on 3D Lidar-based object detection.

\begin{figure}
    \centering
    \includegraphics[width=2in]{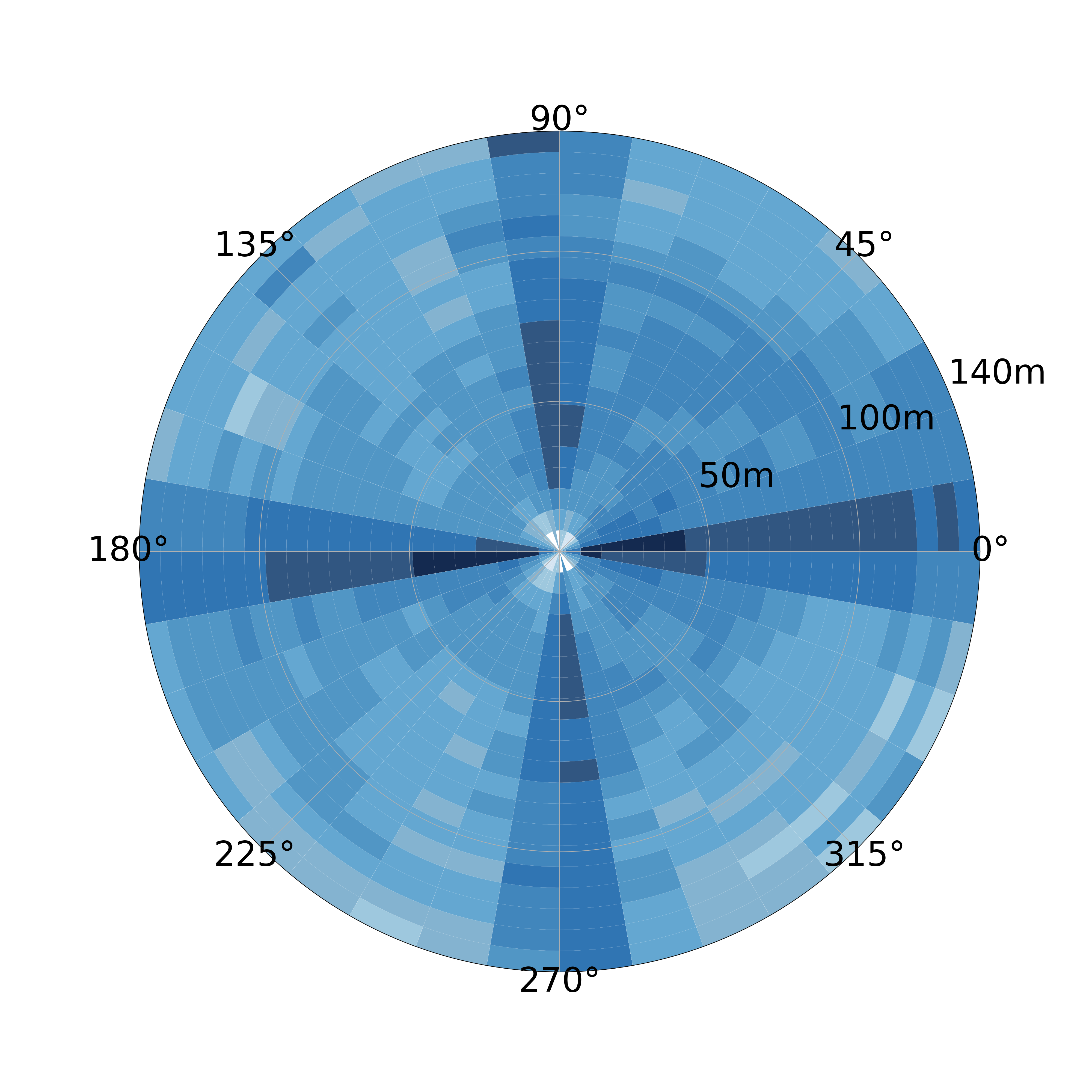}
    \caption{Polar density map in log scale for ground truth bounding boxes. The polar and radial axes indicate the angle and distance (in meters) of the bounding boxes with respect to the ego vehicle. The color indicates the number of bounding boxes~(log scale) in the bin. The darker color means a larger number of boxes in the bin.}
    \label{fig:angle_density_map}
\end{figure}
\begin{figure}
    \centering
    \includegraphics[width=3.4in]{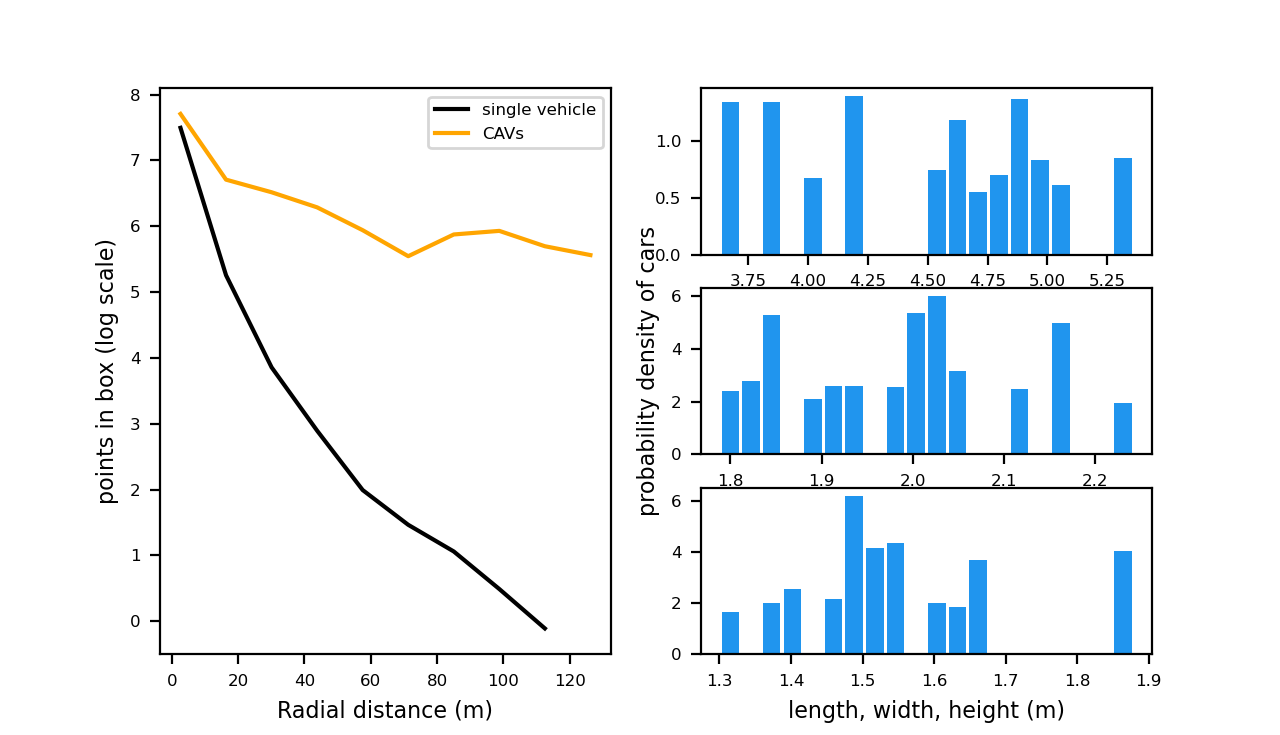}
    \caption{\emph{Left}: Number of points in log scale within the ground truth bounding boxes with respect to radial distance from ego vehicles. \emph{Right}: Bounding box size distributions. }
    \label{fig:bbx_dist}
\end{figure}

\begin{table*}[]
\centering
\caption{Summary of OPV2V dataset statistics. Traffic density means the number of vehicles spawned around the ego vehicle within a 140m radius and aggressiveness represents the probability of a vehicle operating aggressive overtakes.}
\begin{tabular}{l|lllllll}
\toprule
Road Type &
  Percentage(\%) &
  \begin{tabular}[c]{@{}l@{}}Length(s)\\ mean/std\end{tabular} &
  \begin{tabular}[c]{@{}l@{}}CAV number\\   mean/std\end{tabular} &
  \begin{tabular}[c]{@{}l@{}}Traffic density\\      mean/std\end{tabular} &
  \begin{tabular}[c]{@{}l@{}}Traffic Speed(km/h)\\           mean/std\end{tabular} &
  \begin{tabular}[c]{@{}l@{}}CAV speed(km/h)\\           mean/std\end{tabular} &
  \begin{tabular}[c]{@{}l@{}}Aggressiveness\\       mean/std\end{tabular} \\ \hline
4-way Intersection & 24.5 & 12.5/4.2  & 2.69/0.67 & 29.6/26.1 & 19.3/8.8  & 21.3/10.2 & 0.09/0.30 \\
T Intersection     & 24.1 & 14.3/12.8  & 2.55/1.3 & 27.9/18.65  & 26.3/7.5  & 26.2/10.0  & 0.11/0.32 \\
Straight Segment   & 20.7 & 20.2/12.7 & 3.54/1.21 & 38.0/36.3 & 45.7/14.8 & 54.3/20.1 & 0.82/0.40 \\
Curvy Segment      & 23.3 & 17.8/6.8  & 2.86/0.95 & 19.1/9.2  & 45.8/15.1 & 51.6/19.2 & 0.50/0.51 \\
Midblock           & 4.7  & 10.0/1.3  & 3.00/1.22 & 21.8/8.2  & 45.1/8.3  & 50.7/11.5 & 0.20/0.44 \\
Entrance Ramp      & 2.7  & 9.3/0.9   & 2.67/0.57 & 20.3/2.8  & 54.8/1.7  & 66.7/4.8  & 0.67/0.57 \\
Overall            & 100  & 16.4/9.1  & 2.89/1.06 & 26.5/17.2 & 33.1/15.8 & 37.5/21.0 & 0.34/0.47 \\ \bottomrule
\end{tabular}
\label{table:over_view}
\end{table*}

\begin{figure*}[!t]
\centering
\includegraphics[width=0.8\textwidth]{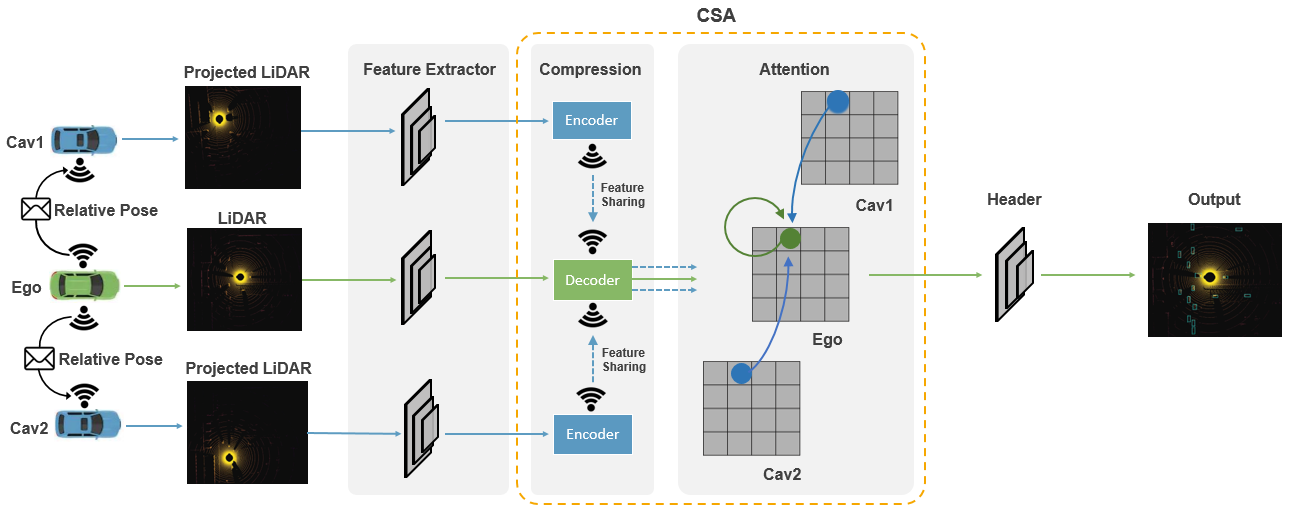}
\caption{The architecture of Attentive Intermediate Fusion pipeline. Our model consists of 6 parts: 1) Metadata Sharing: build connection graph and broadcast locations among neighboring CAVs. 2) Feature Extraction: extract features based on each detector's backbone. 3) Compression (optional): use Encoder-Decoder to compress/decompress features. 4) Feature sharing: share (compressed) features with connected vehicles. 5) Attentive Fusion: leverage self-attention to learn interactions among features in the same spatial location. 6) Prediction Header: generate final object predictions.}
\label{fig:intermediate}
\end{figure*} 

\subsection{Data Analysis}
As Table~\ref{table:over_view} depicts, six distinct categories of road types are included in our dataset for simulating the most common driving scenarios in real life. To minimize data redundancy, we attempt to avoid overlong clips and assign the ego vehicles short travels with an average length of 16.4 seconds,  dissimilar locations,  and divergent maneuvers for each scenario. We also allocate the gathered 73 scenes with diverse traffic and CAV configurations to enlarge dataset variance.

Fig.~\ref{fig:angle_density_map} and Fig.~\ref{fig:bbx_dist} reveal the statistics of the 3D bounding box annotations in our dataset.  Generally, the cars around the ego vehicle are well-distributed with divergent orientations and bounding box sizes. This distribution is in agreement with the data collection process where the object positions are randomly selected around CAVs and vehicle models are also arbitrarily chosen. As shown in Fig.~\ref{fig:angle_density_map}, unlike the dataset for the single self-driving car, our dataset still has a large portion of objects in view with distance $\geq$ 100m, given that the ground truth boxes are defined with respect to the aggregated lidar points from all CAVs. As displayed in Fig.~\ref{fig:bbx_dist}, although a single vehicle's LiDAR points for distant objects are especially sparse, other CAVs are able to provide compensations to remarkably boost the LiDAR points density. This demonstrates the capability of V2V technology to drastically increase perception range and provide compensation for occlusions.

\section{Attentive Intermediate Fusion Pipeline}
As sensor observations from different connected vehicles potentially carry various noise levels (e.g., due to distance between vehicles), a method that can pay attention to important observations while ignoring disrupted ones is crucial for robust detection. Therefore, we propose an Attentive Intermediate Fusion pipeline to capture the interactions between features of neighboring connected vehicles, helping the network attend to key observations. The proposed Attentive Intermediate Fusion pipeline consists of 6 modules: Metadata sharing, Feature Extraction, Compression, Feature sharing, Attentive Fusion, and Prediction. The overall architecture is shown in Fig.~\ref{fig:intermediate}. The proposed pipeline is flexible and can be easily integrated with existing Deep Learning-based LiDAR detectors~(see Table~\ref{table:benchmark_analysis}).

\noindent\textbf{Metadata Sharing and Feature Extraction: }We first broadcast each CAVs' relative pose and extrinsics to build a spatial graph where each node is a CAV within the communication range and each edge represents a communication channel between a pair of nodes. After constructing the graph, an ego vehicle will be selected within the group.\footnote{During training, a random CAV within the group is selected as ego vehicle while in the inference, the ego vehicle is fixed for a fair comparison.} And all the neighboring  CAVs will project their own point clouds to the ego vehicle's LiDAR frame and extract features based on the projected point clouds. The feature extractor here can be the backbones of existing 3D object detectors.

\noindent\textbf{Compression and Feature sharing: }An essential factor in V2V communication is the hardware restriction on transmission bandwidth. The transmission of the original high-dimensional feature maps usually requires large bandwidth and hence compression is necessary.  One key advantage of intermediate fusion over sharing raw point clouds is the marginal accuracy loss after compression~\cite{Wang2020V2VNetVC}. Here we deploy an Encoder-Decoder architecture to compress the shared message. The Encoder is composed of a series of 2D convolutions and max pooling, and the feature maps in the bottleneck will broadcast to the ego vehicle. The Decoder that contains several deconvolution layers~\cite{noh2015learning} on the ego-vehicles' side will recover the compressed information and send it to the Attentive Fusion module. 

\noindent\textbf{Attentive Fusion: }Self-attention models~\cite{vaswani2017attention} are adopted to fuse those decompressed features. Each feature vector (green/blue circles shown in Fig.\ref{fig:intermediate}) within the same feature map corresponds to certain spatial areas in the original point clouds. Thus, simply flattening the feature maps and calculating the weighted sum of features will break spatial correlations. Instead, we construct a local graph for each feature vector in the feature map, where edges are built for feature vectors in the same spatial locations from disparate connected vehicles. One such local graph is shown in Fig.\ref{fig:intermediate} and self-attention will operate on the graph to reason the interactions for better capturing the representative features. 

\noindent\textbf{Prediction Header: }The fused features will be fed to the prediction header to generate bounding box proposals and associated confidence scores. 

\begin{figure}
    \centering
    \includegraphics[width=1.8in]{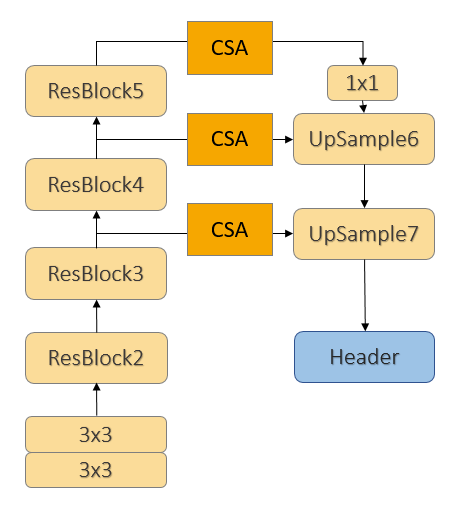}
    \caption{The architecture of PIXOR with Attentive Fusion. }
    \label{fig:pixor}
\end{figure}

\section{Experiments}

\subsection{Benchmark models}
We implement four state-of-the-art LiDAR-based 3D object detectors on our dataset and integrate these detectors with three different fusion strategies i.e., early fusion, late fusion, and intermediate fusion. We also investigate the model performance under a single-vehicle setting, named no fusion, which neglects V2V communication. Therefore, in total 16 models will be evaluated in the benchmark. All the models are implemented in a unified code framework, and  our code and develop tutorial can be found in the \textcolor{blue}{\href{https://mobility-lab.seas.ucla.edu/opv2v/}{project website}}.

\noindent\textbf{Selected 3D Object Detectors: } 
We pick SECOND~\cite{Yan2018SECONDSE}, VoxelNet~\cite{voxelnet}, PIXOR~\cite{Yang2018PIXORR3}, and PointPillar~\cite{pointpillar} as our 3D LiDAR detectors for benchmarking analysis.

\noindent\textbf{Early fusion baseline: }All the LiDAR point clouds will be projected into ego-vehicles' coordinate frame, based on the pose information shared among CAVs, and then the ego vehicle will aggregate all received point clouds and feed them to the detector.

\noindent\textbf{Late fusion baseline: } Each CAV will predict the bounding boxes with confidence scores independently and broadcast these outputs to the ego vehicle. Non-maximum suppression~(NMS) will be applied to these proposals afterwards to generate the final object predictions.

\noindent\textbf{Intermediate fusion: }The Attentive Fusion pipeline is flexible and can be easily generalized to other object detection networks. To evaluate the proposed pipeline, we only need to add the Compression, Sharing, and Attention~(CSA) module to the existing network architecture. Since 4 different detectors add CSA modules in a similar way, here we only show the architecture of intermediate fusion with the PIXOR model as Fig.~\ref{fig:pixor} displays.  Three CSA modules are added at the 2D backbone of PIXOR to aggregate multi-scale features while all other parts of the network remain the same. 

\begin{table}[]
    \centering
    \caption{Object detection results on Default CARLA Towns and digital Culver City.}
    \begin{tabular}{c|c|cc|cc}
        \toprule
        \multicolumn{2}{c|}{\multirow{3}{*}{Method}}&\multicolumn{2}{c|}{Default}&\multicolumn{2}{c}{Culver}\\
        \multicolumn{2}{c|}{}&\multicolumn{2}{c|}{AP@IoU}&\multicolumn{2}{c}{AP@IoU}\\
         \multicolumn{2}{c|}{}&$0.5$&$0.7$&$0.5$&$0.7$ \\
         \hline
        \multirow{4}{*}{PIXOR}&No Fusion&0.635 &0.406 &0.505&0.290\\
        &Late Fusion&0.769&0.578&0.622&0.360\\
        &Early Fusion&0.810&0.678&\textbf{0.734}&\textbf{0.558}\\
        &Intermediate Fusion&\textbf{0.815}&\textbf{0.687}&0.716&0.549\\
        \hline
        \multirow{4}{*}{PointPillar}&No Fusion&0.679&0.602&0.557&0.471\\
        &Late Fusion&0.858&0.781&0.799&0.668\\
        &Early Fusion&0.891&0.800&0.829&0.696\\
        &Intermediate Fusion&\textbf{0.908}&\textbf{0.815}&\textbf{0.854}&\textbf{0.735}\\
        \hline
        \multirow{4}{*}{SECOND}&No Fusion&0.713&0.604&0.646&0.517\\
        &Late Fusion&0.846&0.775&0.808&0.682\\
        &Early Fusion&0.877&0.813&0.821&0.738\\
        &Intermediate Fusion&\textbf{0.893}&\textbf{0.826}&\textbf{0.875} &\textbf{0.760}\\
        \hline
        \multirow{4}{*}{VoxelNet}&No Fusion&0.688&0.526&0.605&0.431\\
        &Late Fusion&0.801&0.738&0.722&0.588\\
        &Early Fusion&0.852&0.758&0.815&0.677\\
        &Intermediate Fusion&\textbf{0.906}&\textbf{0.864}&\textbf{0.854}&\textbf{0.775}\\
        \bottomrule
    \end{tabular}
    \label{table:benchmark_analysis}
\end{table}

\subsection{Metrics}
We select a fixed vehicle as the ego vehicle among all spawned CAVs for each scenario in the test and validation set. Detection performance is evaluated near the ego vehicle in a range of $x\in[-140, 140]m, y\in[-40, 40]m$. Following \cite{Wang2020V2VNetVC}, we set the broadcast range among CAVs to be 70 meters. Sensing messages outside of this communication range will be ignored by the ego vehicle. Average Precisions~(AP) at Intersection-over-Union (IoU) threshold of both 0.5 and 0.7 are adopted to assess different models. Since PIXOR ignores the $z$ coordinates of the bounding box, we compute IoU only on x-y plane to make the comparison fair. For the evaluation targets, we include vehicles that are hit by at least one LiDAR point from any connected vehicle.

\begin{figure}[!t]
\centering
\includegraphics[width=0.37\textwidth]{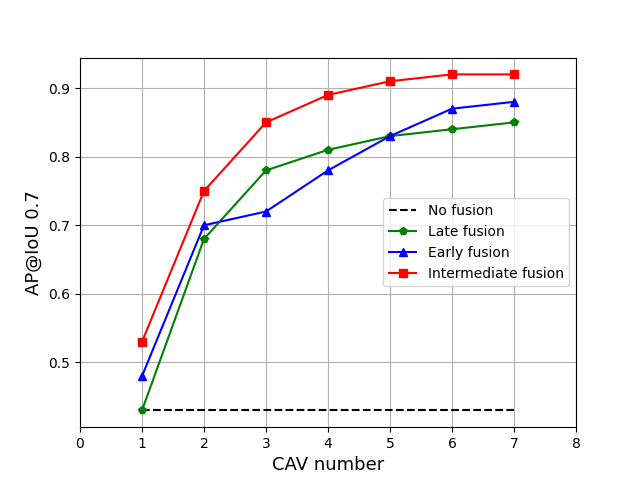}
\caption{Average Precision at IoU=0.7 with respect to CAV number.}
\label{fig:cav_num}
\end{figure}

\begin{figure}[!t]
\centering
\includegraphics[width=0.37\textwidth]{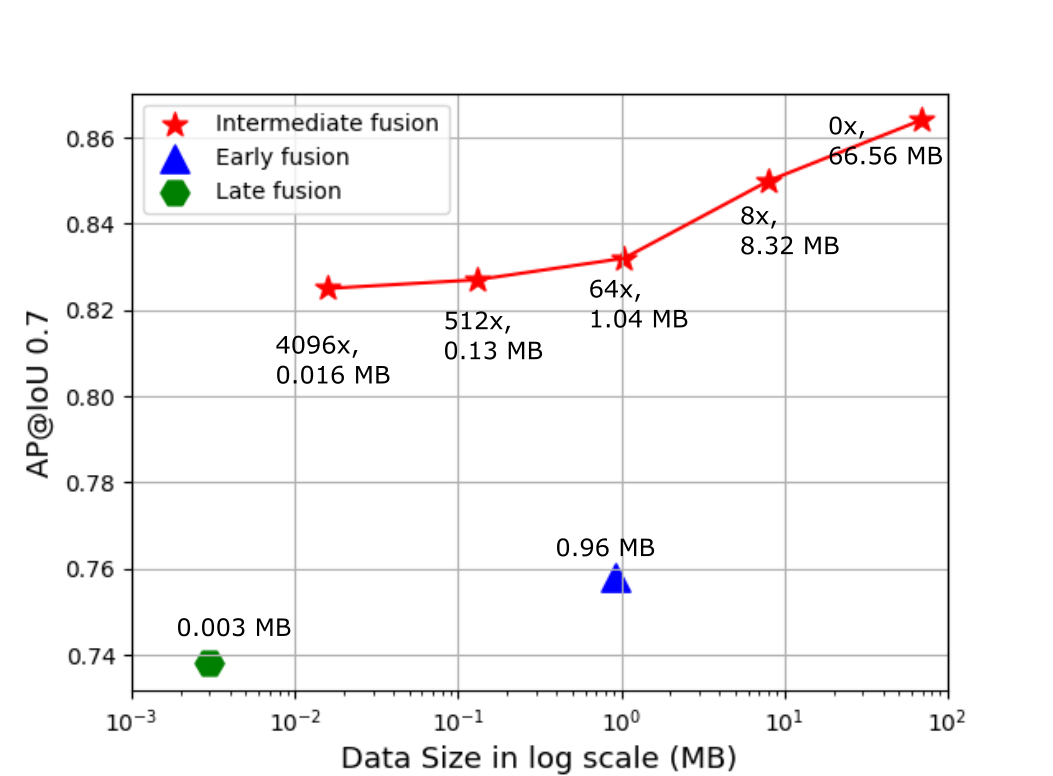}
\caption{Average Precision at IoU=0.7 with respect to data size in log scale based on VoxelNet detector. The number$\times$ refers to the compression rate.}
\label{fig:compression}
\end{figure}
\subsection{Experiment Details}
The train/validation/test splits are 6764/1981/2719 frames. The testing frames contain all road types and are further split into two parts--CARLA default maps and Culver City digital town. For each frame, we assure that the minimum and maximum numbers of CAVs are 2 and 7 respectively. We use Adam Optizer~\cite{kingma2014adam} and early stop to train all models, and it takes us 14 days to finish all training on  4 RTX 3090 GPUs.

\subsection{Benchmark Analysis}
Table~\ref{table:benchmark_analysis} depicts the performance of the selected four LiDAR detectors combined with different fusion strategies. All fusion methods achieve $\geq$10\% AP gains at IoU 0.7 over no fusion counterparts for both default CARLA towns and Culver City, showing the advantage of aggregating information from all CAVs for V2V perception. Generally, because of the capability of preserving more sensing measurements and visual cues, early fusion methods outperform late fusion methods. Except for PIXOR at Culver City, intermediate fusion achieves the best performance on both testing sets compared with all other methods. We argue that the AP gains over early fusion originate from the mechanism of the self-attention module, which can effectively capture the inherent correlation between each CAV's perception information. It is also worth noting that the prediction results for Culver City are generally inferior to CARLA towns. Such a phenomenon is expected as the traffic pattern in Culver City is more similar to real life, which causes a domain gap with the training data. Furthermore, we collect the Culver City data in a busy hour under a very congested driving environment, which leads to vastly severe occlusions and makes the detection task very challenging.

\subsection{Effect of CAV Quantity}
We explore the detection performance as affected by the number of CAVs in a complex intersection scenario where 150 vehicles are spawned in the surrounding area. A portion of them will be transformed into CAVs that can share information. We gradually increase the number of the CAVs up to 7 and apply VoxelNet with different fusion methods for object detection. As shown in Fig.~\ref{fig:cav_num}, the AP has a positive correlation with the number of CAVs. However, when the quantity reaches 4, the increasing rate becomes lower. This can be due to the fact that the CAVs are distributed on different sides of the intersection and four of them can already provide enough viewpoints to cover most of the blind spots. Additional enhancements with 5 or more vehicles come from denser measurements on the same object.

\subsection{Effect of Compression Rates}
Fig.~\ref{fig:compression} exhibits the data size needed for a single transmission between a pair of vehicles and corresponding AP for all fusion methods on the testing set in CARLA towns. We pick VoxelNet for all fusion methods here and simulate distinct compression rates by modifying the number of layers in Encoder-Decoder. By applying a straightforward Encoder-Decoder architecture to squeeze the data, the Attentive Intermediate Fusion obtains an outstanding trade-off between the accuracy and bandwidth. Even with a 4096x compression rate, the performance still just drop marginally~(around 3\%) and surpass the early fusion and late fusion. Based on the V2V communication protocol~\cite{arena2019overview}, data broadcasting can achieve 27 Mbps at the range of 300~m. This represents that the time delay to deliver the message with a 4096x compression rate is only about 5~ms. 

        
        

\section{CONCLUSIONS}
In this paper, we present the first open dataset and benchmark fusion strategies for V2V perception. We further come up with an Attentive Intermediate Fusion pipeline, and the experiments show that the proposed approach can outperform all other fusion methods and achieve state-of-the-art performance even under large compression rates. 

In the future, we plan to extend the dataset with more tasks as well as sensors suites and investigate more multi-modal sensor fusion methods in the V2V and Vehicle-to-infrastructure (V2I) setting. We hope our open-source efforts can make a step forward for the standardizing process of the V2V perception and encourage more researchers to investigate this new direction. 

\addtolength{\textheight}{-3cm}   




\bibliographystyle{IEEEtran}
\bibliography{IEEEfull}

\begin{thebibliography}{10}
\providecommand{\url}[1]{#1}
\csname url@rmstyle\endcsname
\providecommand{\newblock}{\relax}
\providecommand{\bibinfo}[2]{#2}
\providecommand\BIBentrySTDinterwordspacing{\spaceskip=0pt\relax}
\providecommand\BIBentryALTinterwordstretchfactor{4}
\providecommand\BIBentryALTinterwordspacing{\spaceskip=\fontdimen2\font plus
\BIBentryALTinterwordstretchfactor\fontdimen3\font minus
  \fontdimen4\font\relax}
\providecommand\BIBforeignlanguage[2]{{%
\expandafter\ifx\csname l@#1\endcsname\relax
\typeout{** WARNING: IEEEtran.bst: No hyphenation pattern has been}%
\typeout{** loaded for the language `#1'. Using the pattern for}%
\typeout{** the default language instead.}%
\else
\language=\csname l@#1\endcsname
\fi
#2}}

\bibitem{liu2019auto}
C.~Liu, L.-C. Chen, F.~Schroff, H.~Adam, W.~Hua, A.~L. Yuille, and L.~Fei-Fei,
  ``Auto-deeplab: Hierarchical neural architecture search for semantic image
  segmentation,'' in \emph{Proceedings of the IEEE/CVF Conference on Computer
  Vision and Pattern Recognition}, 2019, pp. 82--92.

\bibitem{he2016deep}
K.~He, X.~Zhang, S.~Ren, and J.~Sun, ``Deep residual learning for image
  recognition,'' in \emph{Proceedings of the IEEE conference on computer vision
  and pattern recognition}, 2016, pp. 770--778.

\bibitem{qi2017pointnet}
C.~R. Qi, H.~Su, K.~Mo, and L.~J. Guibas, ``Pointnet: Deep learning on point
  sets for 3d classification and segmentation,'' in \emph{Proceedings of the
  IEEE conference on computer vision and pattern recognition}, 2017, pp.
  652--660.

\bibitem{voxelnet}
Y.~Zhou and O.~Tuzel, ``Voxelnet: End-to-end learning for point cloud based 3d
  object detection,'' 06 2018, pp. 4490--4499.

\bibitem{pointpillar}
A.~Lang, S.~Vora, H.~Caesar, L.~Zhou, J.~Yang, and O.~Beijbom, ``Pointpillars:
  Fast encoders for object detection from point clouds,'' 06 2019, pp.
  12\,689--12\,697.

\bibitem{Shi_2019_CVPR}
S.~Shi, X.~Wang, and H.~Li, ``Pointrcnn: 3d object proposal generation and
  detection from point cloud,'' in \emph{The IEEE Conference on Computer Vision
  and Pattern Recognition (CVPR)}, June 2019.

\bibitem{Shi2020PVRCNNPF}
S.~Shi, C.~Guo, L.~Jiang, Z.~Wang, J.~Shi, X.~Wang, and H.~Li, ``Pv-rcnn:
  Point-voxel feature set abstraction for 3d object detection,'' \emph{2020
  IEEE/CVF Conference on Computer Vision and Pattern Recognition (CVPR)}, pp.
  10\,526--10\,535, 2020.

\bibitem{Liang2009}
M.~Liang, B.~Yang, Y.~Chen, R.~Hu, and R.~Urtasun, ``Multi-task multi-sensor
  fusion for 3d object detection,'' 06 2019, pp. 7337--7345.

\bibitem{Liang2018}
M.~Liang, B.~Yang, S.~Wang, and R.~Urtasun, ``Deep continuous fusion for
  multi-sensor 3d object detection,'' in \emph{Computer Vision -- ECCV 2018},
  V.~Ferrari, M.~Hebert, C.~Sminchisescu, and Y.~Weiss, Eds.\hskip 1em plus
  0.5em minus 0.4em\relax Cham: Springer International Publishing, 2018, pp.
  663--678.

\bibitem{lochrane2020carma}
\BIBentryALTinterwordspacing
T.~Lochrane, L.~Dailey, and C.~Tucker, ``Carma℠: Driving innovation,''
  \emph{Public Roads}, vol.~83, no.~4, 2020. [Online]. Available:
  \url{https://its.dot.gov/cda/}
\BIBentrySTDinterwordspacing

\bibitem{cooper}
\BIBentryALTinterwordspacing
Q.~Chen, S.~Tang, Q.~Yang, and S.~Fu, ``Cooper: Cooperative perception for
  connected autonomous vehicles based on 3d point clouds,'' in \emph{2019 IEEE
  39th International Conference on Distributed Computing Systems
  (ICDCS)}.\hskip 1em plus 0.5em minus 0.4em\relax Los Alamitos, CA, USA: IEEE
  Computer Society, jul 2019, pp. 514--524. [Online]. Available:
  \url{https://doi.ieeecomputersociety.org/10.1109/ICDCS.2019.00058}
\BIBentrySTDinterwordspacing

\bibitem{nuscenes2019}
H.~Caesar, V.~Bankiti, A.~H. Lang, S.~Vora, V.~E. Liong, Q.~Xu, A.~Krishnan,
  Y.~Pan, G.~Baldan, and O.~Beijbom, ``nuscenes: A multimodal dataset for
  autonomous driving,'' \emph{arXiv preprint arXiv:1903.11027}, 2019.

\bibitem{sun2020scalability}
P.~Sun, H.~Kretzschmar, X.~Dotiwalla, A.~Chouard, V.~Patnaik, P.~Tsui, J.~Guo,
  Y.~Zhou, Y.~Chai, B.~Caine, \emph{et~al.}, ``Scalability in perception for
  autonomous driving: Waymo open dataset,'' in \emph{Proceedings of the
  IEEE/CVF Conference on Computer Vision and Pattern Recognition}, 2020, pp.
  2446--2454.

\bibitem{Geiger2013IJRR}
A.~Geiger, P.~Lenz, C.~Stiller, and R.~Urtasun, ``Vision meets robotics: The
  kitti dataset,'' \emph{International Journal of Robotics Research (IJRR)},
  2013.

\bibitem{Wang2020V2VNetVC}
T.-H. Wang, S.~Manivasagam, M.~Liang, B.~Yang, W.~Zeng, J.~Tu, and R.~Urtasun,
  ``V2vnet: Vehicle-to-vehicle communication for joint perception and
  prediction,'' in \emph{ECCV}, 2020.

\bibitem{rawaw018}
Z.~Y. Rawashdeh and Z.~Wang, ``Collaborative automated driving: A machine
  learning-based method to enhance the accuracy of shared information,'' in
  \emph{2018 21st International Conference on Intelligent Transportation
  Systems (ITSC)}, 2018, pp. 3961--3966.

\bibitem{Zhang2021DistributedDM}
Z.~Zhang, S.~Wang, Y.~Hong, L.~Zhou, and Q.~Hao, ``Distributed dynamic map
  fusion via federated learning for intelligent networked vehicles,''
  \emph{ArXiv}, vol. abs/2103.03786, 2021.

\bibitem{xu2021opencda}
R.~Xu, Y.~Guo, X.~Han, X.~Xia, H.~Xiang, and J.~Ma, ``Opencda: An open
  cooperative driving automation framework integrated with co-simulation,'' in
  \emph{2021 IEEE Intelligent Transportation Systems Conference (ITSC)}, 2021.

\bibitem{Dosovitskiy17}
A.~Dosovitskiy, G.~Ros, F.~Codevilla, A.~Lopez, and V.~Koltun, ``{CARLA}: {An}
  open urban driving simulator,'' in \emph{Proceedings of the 1st Annual
  Conference on Robot Learning}, 2017, pp. 1--16.

\bibitem{Rauch2012}
A.~Rauch, F.~Klanner, R.~Rasshofer, and K.~Dietmayer, ``Car2x-based perception
  in a high-level fusion architecture for cooperative perception systems,'' in
  \emph{2012 IEEE Intelligent Vehicles Symposium}, 2012, pp. 270--275.

\bibitem{Rawashdeh2018}
Z.~Y. Rawashdeh and Z.~Wang, ``Collaborative automated driving: A machine
  learning-based method to enhance the accuracy of shared information,'' in
  \emph{2018 21st International Conference on Intelligent Transportation
  Systems (ITSC)}, 2018, pp. 3961--3966.

\bibitem{f-cooper}
\BIBentryALTinterwordspacing
Q.~Chen, X.~Ma, S.~Tang, J.~Guo, Q.~Yang, and S.~Fu, ``F-cooper: Feature based
  cooperative perception for autonomous vehicle edge computing system using 3d
  point clouds,'' in \emph{Proceedings of the 4th ACM/IEEE Symposium on Edge
  Computing}, ser. SEC '19.\hskip 1em plus 0.5em minus 0.4em\relax New York,
  NY, USA: Association for Computing Machinery, 2019, p. 88–100. [Online].
  Available: \url{https://doi.org/10.1145/3318216.3363300}
\BIBentrySTDinterwordspacing

\bibitem{Lidarsim}
S.~Manivasagam, S.~Wang, K.~Wong, W.~Zeng, M.~Sazanovich, S.~Tan, B.~Yang,
  W.-C. Ma, and R.~Urtasun, ``Lidarsim: Realistic lidar simulation by
  leveraging the real world,'' 06 2020, pp. 11\,164--11\,173.

\bibitem{marvasti2020cooperative}
E.~E. Marvasti, A.~Raftari, A.~E. Marvasti, Y.~P. Fallah, R.~Guo, and H.~Lu,
  ``Cooperative lidar object detection via feature sharing in deep networks,''
  in \emph{2020 IEEE 92nd Vehicular Technology Conference
  (VTC2020-Fall)}.\hskip 1em plus 0.5em minus 0.4em\relax IEEE, 2020, pp. 1--7.

\bibitem{OlaverriMonreal2018ConnectionOT}
C.~Olaverri-Monreal, J.~Errea-Moreno, A.~D{\'i}az-{\'A}lvarez, C.~Biurrun-Quel,
  L.~Serrano-Arriezu, and M.~Kuba, ``Connection of the sumo microscopic traffic
  simulator and the unity 3d game engine to evaluate v2x communication-based
  systems,'' \emph{Sensors (Basel, Switzerland)}, vol.~18, 2018.

\bibitem{roadrunner}
\BIBentryALTinterwordspacing
``Roadrunner: Design 3d scenes for automated driving simulation.'' [Online].
  Available: \url{https://www.1stvision.com/cameras/models/Allied-Vision}
\BIBentrySTDinterwordspacing

\bibitem{noh2015learning}
H.~Noh, S.~Hong, and B.~Han, ``Learning deconvolution network for semantic
  segmentation,'' in \emph{Proceedings of the IEEE international conference on
  computer vision}, 2015, pp. 1520--1528.

\bibitem{vaswani2017attention}
A.~Vaswani, N.~Shazeer, N.~Parmar, J.~Uszkoreit, L.~Jones, A.~N. Gomez,
  {\L}.~Kaiser, and I.~Polosukhin, ``Attention is all you need,'' in
  \emph{Advances in neural information processing systems}, 2017, pp.
  5998--6008.

\bibitem{Yan2018SECONDSE}
Y.~Yan, Y.~Mao, and B.~Li, ``Second: Sparsely embedded convolutional
  detection,'' \emph{Sensors (Basel, Switzerland)}, vol.~18, 2018.

\bibitem{Yang2018PIXORR3}
B.~Yang, W.~Luo, and R.~Urtasun, ``Pixor: Real-time 3d object detection from
  point clouds,'' \emph{2018 IEEE/CVF Conference on Computer Vision and Pattern
  Recognition}, pp. 7652--7660, 2018.

\bibitem{kingma2014adam}
D.~P. Kingma and J.~Ba, ``Adam: A method for stochastic optimization,''
  \emph{arXiv preprint arXiv:1412.6980}, 2014.

\bibitem{arena2019overview}
F.~Arena and G.~Pau, ``An overview of vehicular communications,'' \emph{Future
  Internet}, vol.~11, no.~2, p.~27, 2019.

\end{thebibliography}

\end{document}